\documentclass[10pt]{article}

\usepackage{amsmath,amsbsy,amsfonts,amssymb,amsthm,dsfont,fullpage,units}
\usepackage{xspace}
\usepackage{graphicx}
\usepackage{algorithm,algorithmic,mathtools}
\usepackage{color,cases}
\usepackage{subfigure}
\usepackage{multirow}
\usepackage{makecell}
\usepackage{rotating}
\usepackage{bbding}
\usepackage{tikz}

\usepackage{url}
\usepackage{amsfonts}
\usepackage{nicefrac}
\usepackage{microtype}
\usepackage{graphicx}

\usepackage{sidecap}

\def\ie{\emph{i.e.~}}
\def\eg{\emph{e.g.~}}
\def\etal{{\em et al.~}}
\def\aka{{\em a.k.a~}}

\usepackage{hyperref}
\hypersetup{
colorlinks = true,
citecolor = {blue},
citebordercolor = {white}
}

\makeatletter

\title{Saliency Prediction in the Deep Learning Era}

\title{Bottom-up Attention, Models of\footnote{To appear in Encyclopedia of Computational Neuroscience.}}

\author{
Ali Borji\footnote{MarkableAI Inc., Corresponding author}, 
Hamed R. Tavakoli\footnote{Department of Computer Science, Aalto University}, and
Zoya Bylinskii\footnote{Massachusetts Institute of Technology} \\
  \texttt{aliborji@gmail.com, hamed.r-tavakoli@aalto.fi,zoya@mit.edu 
  } \\
}

\begin{document}

\maketitle


\section{Definition}

%

\noindent Attention - a general concept covering all factors that influence selection mechanisms, whether they are scene-driven and bottom-up, or expectation-driven and top-down. \\

\noindent Salience - parts of a stimulus (\eg image, video, audio) that appear to an observer to stand out, relative to their neighboring parts. It is a subjective perceptual quality. \\

\noindent  Gaze - a coordinated motion of the eyes and the head that is long enough to concentrate on something. It is a key property of attention in natural behavior.\\


\noindent  Scene free-viewing - A task in which participants are asked to look at an image, without any specific instruction. As a default scene analysis task, the free viewing paradigm offers a wealth of insights regarding the cues that attract attention.

%
%
%
%
%

\section{Detailed Description}

\subsection{History, scope, and organization}


Deciphering the computational mechanisms by which the brain deals with the computational complexity of an overwhelmingly high volume of incoming data (at a rate of $10^8$ \textit{bits/s}~\cite{KOCH20061428}) and how the brain programs eye movements continue to be important problems in neuroscience. Where humans look in images and videos provides important clues regarding how they perceive static (still images) and dynamic scenes (videos), locate the main focus of the image, recognize actions or events, and identify the main participants. 

A significant amount of behavioral and computational research on attention has revealed that attention is deployed in two ways: bottom-up (BU) and top-down (TD). The bottom-up component of attention, \aka endogenous or stimulus-driven, processes sensory information primarily in a feed-forward manner. Typically, a series of 
successive transformations are applied to the input visual signal to highlight the most interesting, important, conspicuous, or so-called {\em salient} regions~\cite{Koch_Ullman85,Itti_Koch01nrn}. In contrast, in top-down attention, \aka context-driven, or goal-driven, information related to the ongoing behavior, task, or goal is selected (\eg staying within the road lanes while driving~\cite{land1994we}). The reader is referred to~\cite{land2001ways,Ballard1995,hayhoe2005eye,Navalpakkam_Itti05vr,Borji_etal14smc} for reviews of top-down attention studies. Please see also the chapter on top-down attention in this encyclopedia~\cite{schwedhelm2015attentional}.



Conventionally, bottom-up attention models generate a 2D topographic saliency map, where a value at every location determines how salient that location is, relative to its neighbors. The goal in saliency modeling is then to transform an image into its spatially corresponding saliency map (static saliency), possibly also taking into account temporal relations between successive video frames of a movie (dynamic saliency) \cite{Itti_etal98pami}. Early computational saliency models were primarily concerned with identifying conspicuous regions due to low-level feature contrast (\eg pop-out search). Gradually, however, saliency models have shifted from identifying low-level conspicuous image regions, regardless of where people look (as was the goal in Itti~\etal~\cite{Itti_etal98pami}), to predicting eye movements as a proxy for visual saliency (as was the goal in Bruce and Tsotsos~\etal~\cite{bruce2005saliency}). Notice that eye movements are only one way of defining the saliency concept and testing saliency models. There is, however, a subtle distinction between the two. Please see Borji~\etal~\cite{borji2013stands} for a detailed discussion on this. Nonetheless, the recent trend in modeling bottom-up attention and saliency has encouraged researchers to collect increasingly larger eye movement datasets, in particular during free-viewing of natural scenes. 





\begin{figure}[t]
\centering
        \includegraphics[width=\linewidth]{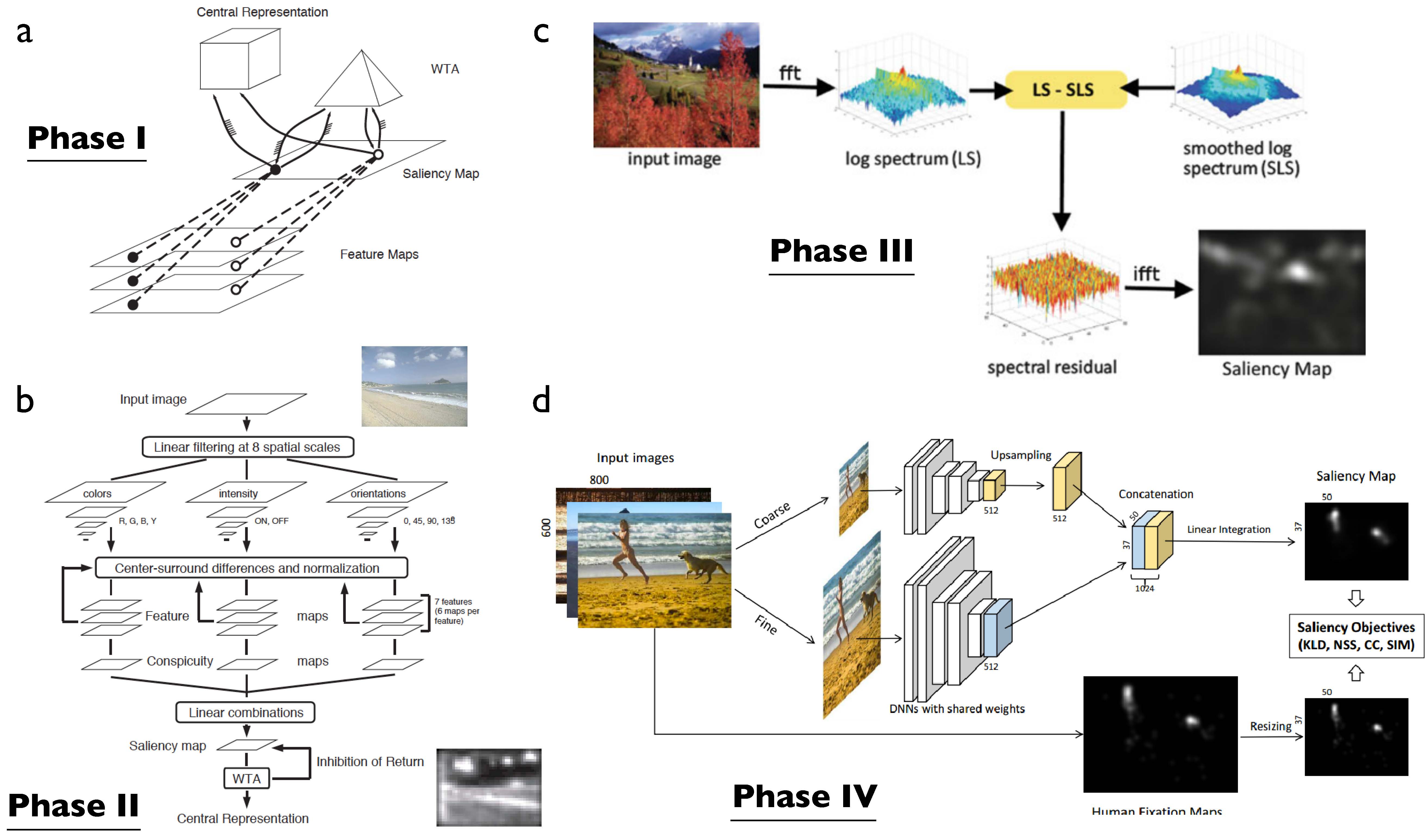}
	\caption{Four models from four different eras of saliency modeling. (a) {\bf Phase I:} Koch \& Ullman (1985) introduced the concept of a saliency map derived from bottom-up inputs from all feature maps,
where a winner-take-all (WTA) network selects the most salient location for further processing. 
(b) {\bf Phase II:} Itti~\etal (1998) proposed a complete computational implementation of a purely
bottom-up and task-independent model based on Koch \& Ullman's theory, including multiscale feature maps,
saliency map, winner-take-all, and inhibition of return.
(c) {\bf Phase III:} A myriad of saliency models appeared between 1998 to 2013. Schematic representation of spectral residual model~\cite{hou2007saliency} by Hou \& Zhang (2007). The log spectrum $\mathcal{L}(f)$ is computed from the down-sampled image (with amplitude $\mathcal{A}(f)$ and
phase $\mathcal{P}(f)$). From $\mathcal{L}(f)$, the spectral residual $\mathcal{R}(f)$ is obtained by multiplying $\mathcal{L}(f)$ with a local average filter and subtracting the result from itself. The saliency map is then the inverse Fourier transform of the exponential of amplitude plus phase (\ie $\mathcal{S}(x) = \mathcal{F}^{-1}\big[exp \big(\mathcal{R}(f) +\mathcal{P}(f) \big)\big]$). 
(d) {\bf Phase IV:} A new wave of saliency models has emerged with the resurgence of convolutional neural networks (CNNs). Huang \etal~\cite{huang2015salicon} proposed a deep saliency model, known as the SALICON, that combines information from two pre-trained CNNs, each on a different image scale (fine and coarse). The two CNNs are then concatenated to produce the final saliency map. } 
	\label{models}
\end{figure}

To frame the concepts and models of bottom-up attention so far into a broader picture, we refer to Figure~\ref{models} as a possible anchor to help organize this review. The computational modeling effort will be reviewed in four phases:

\begin{itemize}

\item {\bf Phase I:} This phase regards the early computational works (\eg Koch and Ullman in 1985; Figure~\ref{models}.a) closely built on top of behavioral hypotheses (\eg the Feature Integration Theory by Triesman \& Gelade). 

\item {\bf Phase II:} Models in this phase extend and operationalize Phase I models to be applicable to an unconstrained variety of stimuli (\eg Itti~\etal model in 1998; Figure~\ref{models}.b). The Itti~\etal model spurred a lot of ideas resulting in a myriad of saliency models in phase III. 

\item {\bf Phase III:} The spectral residual model by Hou and Zhang in 2007 is an example of models in this phase (see Figure~\ref{models}.c for details on this model). It is a simple model and can be implemented in few lines of code.

\item {\bf Phase IV:} Recently, the resurgence and success of neural networks (NN) in computer vision and other areas has brought along a new wave of highly predictive saliency models (Phase IV; Figure~\ref{models}.d). A notable example here is the SALICON model~\cite{huang2015salicon}, proposed in 2015, in conjunction with the SALICON dataset~\cite{jiang2015salicon}. It is the first model to be trained on a large scale attention dataset.
    
\end{itemize}

The principal emphasis in this chapter will be on computational models that can process any visual stimulus in the form of a still image and return a prediction map, the same size as the input image, that can be compared to human or animal behavioral or physiological responses (typically fixations during a task such as free viewing in the context of bottom-up attention)~\cite{rothenstein2008attention}. In addition to these models, some other types of models including abstract models, phenomenological models, or models specifically designed for a single task or for a restricted class of stimuli also exist, but will not be covered here (see~\cite{Itti_Koch01nrn,rothenstein2008attention,gottlieb2010attention,tatler2011eye}).

Bottom-up attention models carry value for at least two main purposes.
First, they present testable predictions that can be utilized for understanding human attention mechanisms at computational, behavioral, and neural levels. Indeed, a large number of cognitive studies have utilized saliency models for model-based hypothesis testing (\eg~\cite{parkhurst2002modeling,Borji_etal13vr}). Second, predicting where people look in images and videos is useful in a wide variety of applications across several domains (\eg computer vision, robotics, neuroscience, medicine, assistive systems, healthcare, and human-computer interaction). Some example applications include gaze-aware compression and summarization, image enhancement, activity recognition, object segmentation, recognition and detection, image captioning, visual question answering, advertisement design, novice training, patient diagnosis, and surveillance. See~\cite{borji2013state} for a review. 

In what follows, we first examine key concepts of early bottom-up attention models
(Section~\ref{classicmodels}), followed by a brief overview of deep saliency models (Section~\ref{deepmodels}), and a discussion of biological plausibility of deep and non-deep saliency models (Section~\ref{bio}). In Section~\ref{benchmarks}, current saliency benchmarks, datasets and new methodologies for collecting large scale data are explained. In Section~\ref{sta}, we provide a quantitative comparison of a large number of deep and non-deep saliency models in their ability to predict eye movements during free viewing of natural scenes. Section~\ref{missing}, explores what current models are missing. Finally, in Section~\ref{conclusion}, we discuss the remaining challenges that need to be addressed in order to build better saliency models.

%
%



\subsection{Bottom-up attention modeling: pre deep learning era}
\label{classicmodels}

Computational modeling of bottom-up attention dates back to the seminal theoretical works by Treisman and Gelade~\cite{treisman1980feature}, the computational architecture by Koch and Ullman~\cite{Koch_Ullman85}, and the bottom-up model of Itti \etal~\cite{Itti_etal98pami}. Itti {\em et al.}'s model was able to predict human behavior in visual search tasks (\eg pop-out versus conjunctive
search \cite{Itti_Koch00vr}), demonstrate robustness to image noise \cite{Itti_etal98pami},
detect traffic signs and other salient objects in natural environments \cite{Itti_Koch01ei}, detect pedestrians in natural scenes \cite{Miau_etal01spie}, locate military vehicles in overhead imagery \cite{Itti_etal01oe}, and~---~most importantly~---~predict where humans look during passive viewing of images and videos \cite{Parkhurst_etal02,Peters_etal05vr}. Note that 
visual salience does not only depend on the physical property of a visual stimulus. It is a consequence of the interaction of a stimulus with other stimuli, as well as with a visual system (biological or
artificial). For example, a color-blind person may have a dramatically different experience of visual salience than a
person with normal color vision. 

Following initial success, many research groups started exploring the notions of bottom-up attention and visual
salience, which gave rise to many computational models from 1998 to 2013. In 2013, we summarized 53 bottom-up models along 13 different factors~\cite{borji2013state,borji2013quantitative}. These models fall into different categories (\eg Bayesian, learning-based, spectral, cognitive). The early models mainly computed visual salience from bottom-up features in several feature maps, including luminance contrast, red-green and blue-yellow color opponency, and oriented edges \cite{Itti_etal98pami}. Subsequent models incorporated mid- and higher level features (\eg face and text~\cite{2009Cerf}, gaze direction~\cite{Parks_etal15vr}) to better predict gaze. A thorough examination of all models in this period is certainly not feasible in this limited space. Instead, we list a number of highly influential static and dynamic saliency models as follows. These include both models that are strongly inspired by biological vision, as well as other implementations of saliency that are based on more abstract mathematical definitions.

%

\begin{enumerate}
\item {\bf Static saliency models:} Attention for Information Maximization (AIM)~\cite{bruce2005saliency}, Graph-based Visual Saliency (GBVS)~\cite{harel2006graph}, Saliency Using Natural statistics (SUN)~\cite{zhang2008sun}, Spectral Residual saliency (SR)~\cite{hou2007saliency}, Adaptive Whitening Saliency (AWS)~\cite{garcia2012saliency}, Boolean Map based Saliency (BMS)~\cite{zhang2013saliency}, and the Judd~\etal model~\cite{judd2009learning}. 

\item {\bf Dynamic saliency models:} AWS-D~\cite{leboran2017dynamic}, OBDL~\cite{hossein2015many}, Xu \etal~\cite{xu2017learning}, PQFT~\cite{guo2008spatio}, and Rudoy \etal~\cite{rudoy2013learning}. 

\end{enumerate}


\subsection{Bottom-up attention modeling: deep learning era}
\label{deepmodels}

Deep learning has emerged as a very successful solution to a variety of problems across several computational domains~\cite{LeCun_Cortes}.  
A deep-learning architecture is a cascade of simple modules that compute non-linear input-output mappings, and all (or most) of which are subject to learning. A certain type of deep architectures, known as convolutional neural networks (CNN, \aka ConvNets) has been very popular. A typical CNN is composed of a series of convolutional layers and pooling layers, followed by one or more fully connected layers. The parameters of the entire network are learned via backpropagation over a large scale labeled dataset for a certain task (\eg object recognition). 
The overall architecture of CNNs resembles the LGN-V1-V2-V4-IT hierarchy of the visual ventral stream, and 
the convolutional and pooling layers are directly inspired by the classic notions of simple cells and complex cells in the cortex (\eg~\cite{yamins2016using}).

The success of CNNs on large scale object recognition corpuses~\cite{deng2009imagenet}, has brought along a new wave of saliency models that perform markedly better than 
traditional saliency models based on hand-crafted features. 
To model bottom-up attention, researchers leverage existing deep architectures that are trained for scene or object recognition and re-purpose them to predict saliency. Often some architectural novelties are also introduced. These models are trained in an end-to-end manner, effectively formulating saliency as a regression problem. To remedy the lack of sufficiently large scale fixation datasets, deep saliency models are often pre-trained on large image datasets and are then fine-tuned on small scale eye movement or click datasets. This procedure allows models to re-use the object-level visual knowledge already learned in CNNs and successfully transfer them to the task of saliency prediction. A large number of deep saliency models have appeared in a relatively short period of time (2014-2018). A detailed discussion of these models goes beyond the scope of this chapter. Instead, we include a number of landmark static and dynamic deep saliency models and refer the reader to~\cite{borjiNew}, for a comprehensive review. These models differ in their architectures and the way they are trained. 


\begin{enumerate}
\item {\bf Static saliency models:} eDN~\cite{vig2014large}, DeepGaze I \& II~\cite{kummerer2014deep}, Mr-CNN~\cite{liu2015predicting}, SALICON~\cite{huang2015salicon}, DeepFix~\cite{kruthiventi2017deepfix}, SAM-ResNet~\cite{cornia2016predicting}, and EML-Net~\cite{jia2018eml}.

\item {\bf Dynamic saliency models:} Two-stream network~\cite{bak2018spatio}, Chaabouni \etal~\cite{chaabouni2016transfer}, Bazzani \etal~\cite{bazzani2016recurrent}, OM-CNN~\cite{jiang2017predicting}, Gorji \& Clark~\cite{gorji2018going}, ACLNet~\cite{wang2018revisiting}, and SG-FCN~\cite{sun2018sg}.

\end{enumerate}

To account for differences in viewing static and dynamic stimuli by human observers\footnote{Observers view images and videos differently. They have much less time to view each video frame (about 1/30 of a second) compared to 3 to 5 seconds over still images. Further, motion is a key component that is missing in still images but strongly attracts human attention over videos (See~\cite{rudoy2013learning}).},
traditional video saliency models pair bottom-up feature extraction
with an ad-hoc motion estimation method that can be performed either by means of optical flow or feature tracking. In contrast, deep video saliency models learn the entire process end-to-end, either by adding temporal information to CNNs (as in~\cite{bak2018spatio}), or developing a dynamic structure using recurrent neural networks~\cite{Hochreiter_Schmidhuber97} (as in~\cite{bazzani2016recurrent}).


\subsection{Biological plausibility of classic and deep saliency models} 
\label{bio}
How well do classic and deep saliency models agree with biological findings on visual attention mechanisms?
To answer this question, we would like to highlight two key points. First, as explained above, previous research has shown that CNNs can explain the feed-forward mechanisms involved in rapid object recognition~\cite{yamins2016using}. Second, since deep saliency models are built on top of CNNs, they inherit biologically-plausible properties of CNNs (\eg convolution operation). While it is not entirely clear how saliency is computed in these models and whether they work similar to traditional saliency models (\eg by implementing center-surround operations, normalization, etc; see Figure~\ref{fig:bruce}), there is evidence that they may generalize classic models. To get an idea, consider a CNN with a single convolutional layer followed by a fully connected layer trained to predict fixations. This model generalizes the classic Itti model and also models built upon it that learn to combine feature maps (\eg~\cite{Judd2009,borji2012boosting,xu2014predicting}). The learned features in the CNN will correspond to orientation, color, intensity, etc. which can be combined linearly by a fully connected layer (or 1x1 convolutions in a fully convolutional neural network). To handle the scale dependency of saliency computation, classic models often utilize multiple image resolutions. In addition to this technique (as in the SALICON model), deep saliency models concatenate maps from several convolutional layers (as in ML-Net), or 
combine input from earlier layers in the network with later layers (\eg using skip connections) to preserve fine details. 


Despite the above resemblances, the most evident shortcoming of classic models (\eg the Itti model) with respect to today's deep architectures is the lack of ability to extract higher level features, objects, or parts of objects. Some classic models remedied this shortcoming by explicitly incorporating object detectors such as face or text detectors.
The hierarchical deep structure of CNNs (\eg 152 layers in the ResNet~\cite{he2016deep}) allows capturing complex cues that attract gaze automatically. This is perhaps the main reason behind the big performance gap between the two types of models. In practice, however, research has shown that there are cases where classic saliency models win over the deep models~\cite{huang2015salicon}, indicating that current deep models still fall short in fully explaining low-level saliency (See Figure~\ref{fig:bruce}). Further, deep models still fail in capturing some high-level attention cues (\eg gaze direction, objects of action, relative importance of objects; See Figure~\ref{fig:failures}). 

Our understanding of how saliency computation emerges inside deep saliency architectures, and how the mechanisms involved in these models differ from those implemented in deep models for object recognition are still limited. In this regard, recent work on understanding the representations learned by CNNs for scene and object recognition can offer new insights to understand deep saliency models (\eg~\cite{zhou2014object,ourVis}).

%
%



\subsection{Benchmarks, datasets, and new data collection methodologies}
\label{benchmarks}

Benchmarks have been instrumental for advances in computer vision~\cite{borjiNew}. In the saliency domain, they have sparked a lot of interest and have spurred a lot of interesting ideas over the past several years. Two of the most influential image-based benchmarks include MIT\footnote{\href{saliency.mit.edu}{saliency.mit.edu}} and SALICON\footnote{\href{http://salicon.net}{http://salicon.net}}. 

The MIT benchmark is currently the gold standard for evaluating and comparing image-based saliency models. It supports eight evaluation measures for comparison and reports results over two eye movement datasets: 1) MIT300 and 2) CAT2000~\cite{Borji_Itti15arxiv}. As of October 2018, 85 models are evaluated over the MIT300 dataset, out of which 26 are NN-based models ($\sim$30\% of all submissions). The CAT2000 dataset has 30 models evaluated to date (9 are NN-based). In addition, 5 baselines are computed on both datasets. The SALICON benchmark is relatively new and is primarily based on the SALICON dataset~\cite{huang2015salicon}. It offers results over 7 scores and uses the same evaluation tools as the MIT benchmark. These two benchmarks are complementary to each other. The former evaluates models with respect to actual fixations but suffers from small scale data. The latter fixes the scale problem but considers noisy click data as a proxy of attention. In this review, we provide an overview of the SALICON dataset, since it has been very useful for constructing saliency models, but focus on providing results over the MIT benchmark since fixations provide a closer link to visual attention mechanisms than mouse clicks~\cite{tavakoli2017saliency}. 

\begin{figure}[t]
	\centering
    \includegraphics[width=\linewidth]{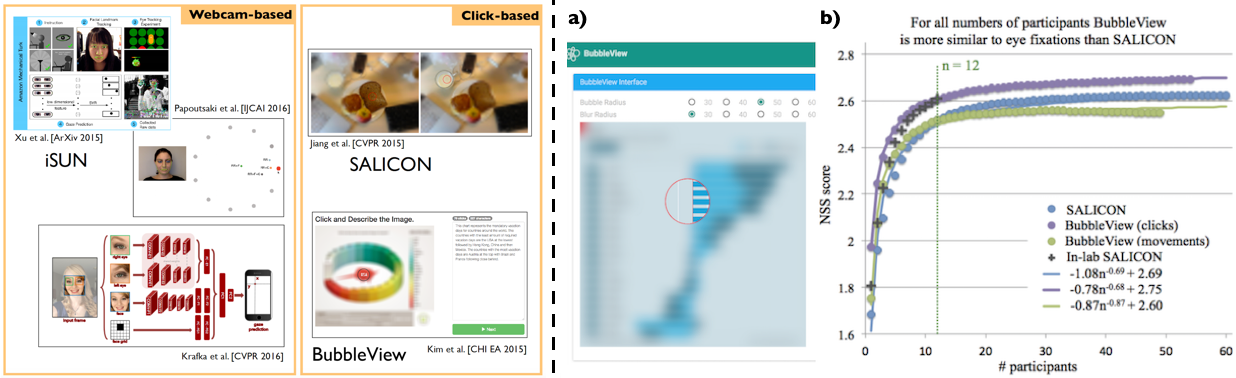} 
	\caption{Left: New crowd-sourcing methodologies to collect attention data including TurkerGaze~\cite{xu2015turkergaze}, SALICON~\cite{jiang2015salicon}, and BubbleView~\cite{kim2017bubbleview}.
	Right: a) An illustration of BubbleView~\cite{kim2017bubbleview} paradigm, b): 
	The NSS score obtained by comparing mouse clicks and mouse movements to ground truth fixations on natural images on the OSIE dataset~\cite{xu2014predicting}. Each point represents the score obtained for a given number of participants, averaged over 10 random splits of participants and all 51 images used. It shows that BubbleView clicks better approximate fixations than SALICON mouse movements for all feasible numbers of participants ($n < 60$). It also shows that the clicks of 10 participants explain 90\% of fixations. Figure compiled from Kim~\etal~\cite{kim2017bubbleview}.}
	\label{fig:header2}
	    \vspace*{-10pt}
\end{figure}

Traditionally, saliency models have been validated by comparing their outputs on small scale datasets composed of eye movements of humans or monkeys watching complex image or video stimuli (\eg~\cite{parkhurst2002modeling,bruce2005saliency}). New large scale databases have emerged by following two trends, 1) increasing the number of images, and 2) introducing new measurements to saliency by providing contextual annotations (\eg image categories, regional properties, etc.). To annotate these large scale datasets, researchers have resorted to crowd-sourcing schemes such as gaze tracking using webcams~\cite{xu2015turkergaze} or mouse movements~\cite{jiang2015salicon,kim2017bubbleview} as alternatives to lab-based eye trackers (Figure~\ref{fig:header2}). Deep supervised saliency models rely heavily on these sufficiently large and well-labeled datasets. Here, we provide an overview of the most recent and popular image datasets for training and testing saliency models. For a review of fixation datasets pre-deep learning era please consult~\cite{winkler2013overview}.


\begin{itemize}

\item \textbf{MIT300:} This dataset is a collection of 300 natural images from the Flickr Creative Commons and personal
collections~\cite{mit-saliency-benchmark}. It contains eye movement data of 39 observers which results in a fairly robust ground-truth to test models against. It is a challenging dataset for saliency models, as images are highly
varied and natural. Fixation maps of all images are held out and used by the MIT Saliency Benchmark for evaluating models.

\item \textbf{CAT2000:} Released in 2015, this is a relatively larger dataset consisting of 2000 training images and 2000 test images spanning 20 different categories such as Cartoons, Art, Satellite, Low resolution images, Indoor, Outdoor, Line drawings, etc.~\cite{mit-saliency-benchmark}. Images in this dataset come from search engines and computer vision
datasets. The training set contains 100 images per category and has fixation annotations from 18 different observers. The test test, used for evaluation, contains the fixations of 24 observers. Both MIT300 and CAT2000 datasets are collected using the EyeLink1000 eye-tracker.

\item \textbf{SALICON:} It is currently the largest crowd-sourced saliency dataset. Images of this dataset come from the Microsoft COCO dataset and contain pixelwise semantic annotations. The SALICON dataset contains 10,000 training images, 5,000 validation images and 5,000 test images. 
Mouse movements are collected using Amazon Mechanical Turk via a psychophysical paradigm known as \textit{mouse-contingent saliency annotation} (Figure~\ref{fig:header2}). Eye movements sometimes do not match mouse movements~\cite{tavakoli2017saliency}. Nevertheless, this dataset introduces an acceptable and scalable method for the collection of additional data for saliency modeling. Currently, many deep saliency models are first trained on the SALICON dataset and are then finetuned on the MIT1000 or CAT2000 datasets for predicting fixations. A similar paradigm known as the BubbleView~\cite{kim2017bubbleview} has also been proposed where a subject has to successively click on a blurred image to reveal the story of the scene.

\end{itemize}

%


\subsection{State of the art performance}
\label{sta}

A quantitative comparison of static saliency models over the MIT benchmark is presented here\footnote{For detailed results please refer to~\cite{borjiNew}. Since our concentration is on the ability of models to predict saliency measured via eye movements, we do not report results on the SALICON dataset.}. 
The MIT benchmark has the most comprehensive set of traditional and deep saliency models evaluated over eight scores. We mainly focus on performances using the AUC-Judd and NSS scores since they provide a better model assessment than others~\cite{bylinskii2018different}. The following 5 baselines are also considered:





\begin{enumerate}
\item {\bf Infinite humans:} How well a fixation map of infinite observers predicts fixations from a different set of infinite observers, computed as a limit? See~\cite{bylinskii2018different} for details. Prediction is still not perfect due to observer differences.

\item {\bf One human:} How well a fixation map of one observer (taken as a saliency map) predicts the fixations of the other $n-1$ observers. This is computed for each observer in turn, and averaged over all $n$ observers. Different individuals are more or less predictive of the rest of the population, and so a range of prediction scores is obtained. 


\item {\bf Center:} This saliency model is computed by stretching a symmetric Gaussian to fit the aspect ratio of a given image, under the assumption that the center of the image is most salient~\cite{tatler2007central}.

\item {\bf Permutation control:} For each image, instead of randomly sampling fixations, fixations from a randomly-sampled image are chosen as the saliency map. This process is repeated 5 times per image, and the average  performance is computed. This method allows capturing observer and center biases that are independent of the image 
\cite{koehler2014saliency}.

\item {\bf Chance:} A random uniform value is assigned to each image pixel to build a saliency map. Average performance is computed over 5 such chance saliency maps per image. 

\end{enumerate}

\subsubsection{Model comparison over the MIT300 dataset} 
\begin{figure}
	\centering
    \includegraphics[width=\linewidth ]{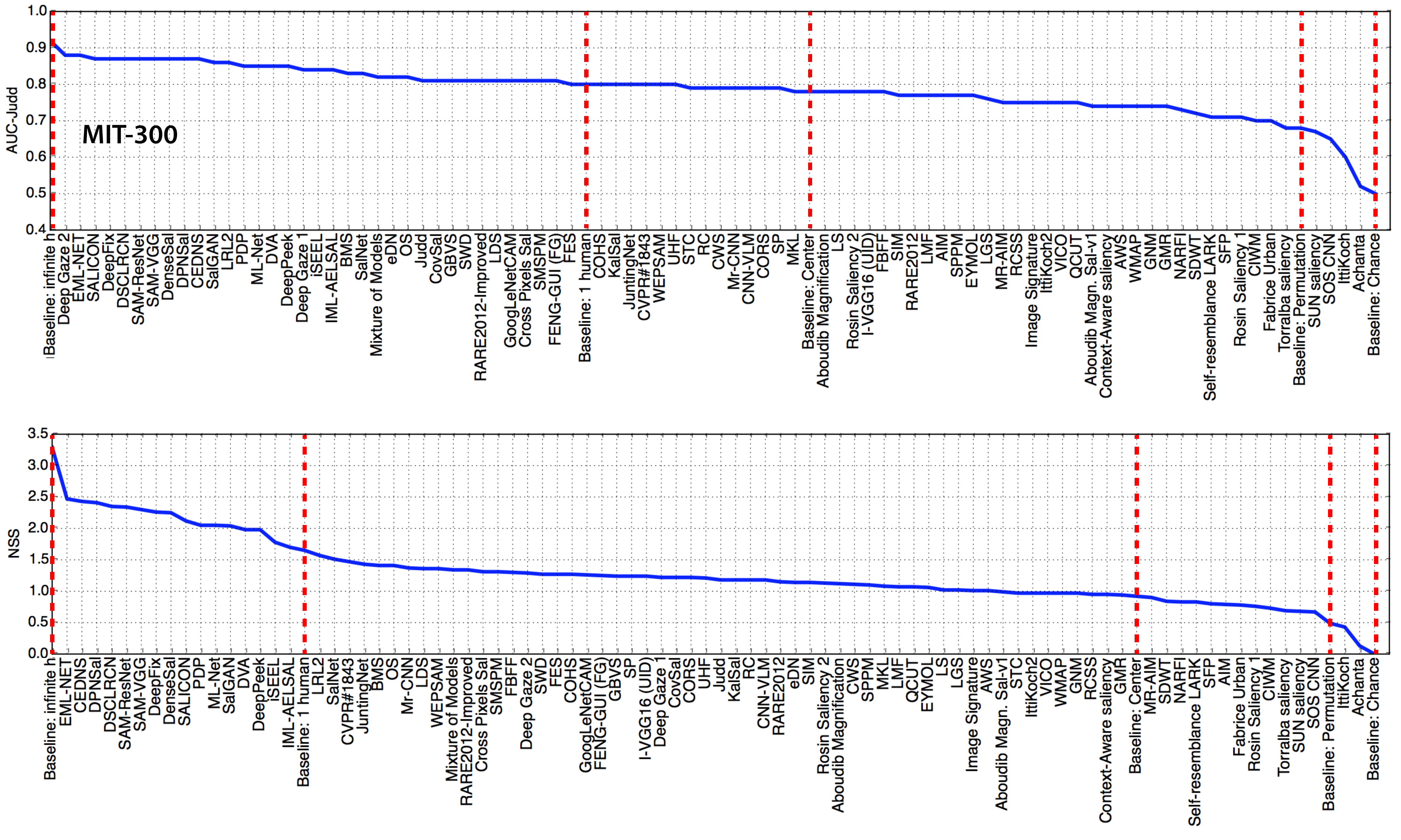} \\
	\caption{Performance of 85 saliency models and 5 baselines (marked with dashed red lines) for fixation prediction over the MIT300 dataset. Models are sorted in terms of the AUC-Judd (top row) and NSS (bottom row) scores. Infinite h. stands for \textit{infinite humans} baseline.}
	\label{fig:MitRes}
\end{figure}

Figure~\ref{fig:MitRes} shows the model performance over the MIT300 dataset. According to the AUC-Judd and NSS scores, the top 5 models are all NN-based. Using the AUC-Judd measure, DeepGaze II and EML-NET models hold the top 2 spots. The \textit{infinite humans} baseline performs the best.
While models perform close to each other according to the AUC-Judd score, switching to NSS widens the differences. EML-NET has the highest score. The \textit{infinite humans} baseline achieves the best NSS. The second and third ranks here belong to CEDNS and DPNSal models. 
BMS is the best-performing non NN-based model and ranks better than the deep eDN model. The majority of models significantly outperform all the baselines, except the \textit{infinite humans} baseline. Among baselines, \textit{one human} and \textit{center} predict fixations better than \textit{permutation} and \textit{chance}. For details please see~\cite{borjiNew}.



Majority of the top 30 models are NN-based. Comparing the best results pre and during the deep learning era shows significant improvements in terms of NSS and AUC scores. At the same time, the gap between the best model and the \textit{infinite humans} baseline has also reduced. 


%
%

\subsubsection{Model comparison over the CAT200 dataset}
\begin{figure}
	\centering
    \includegraphics[width=.6\linewidth ]{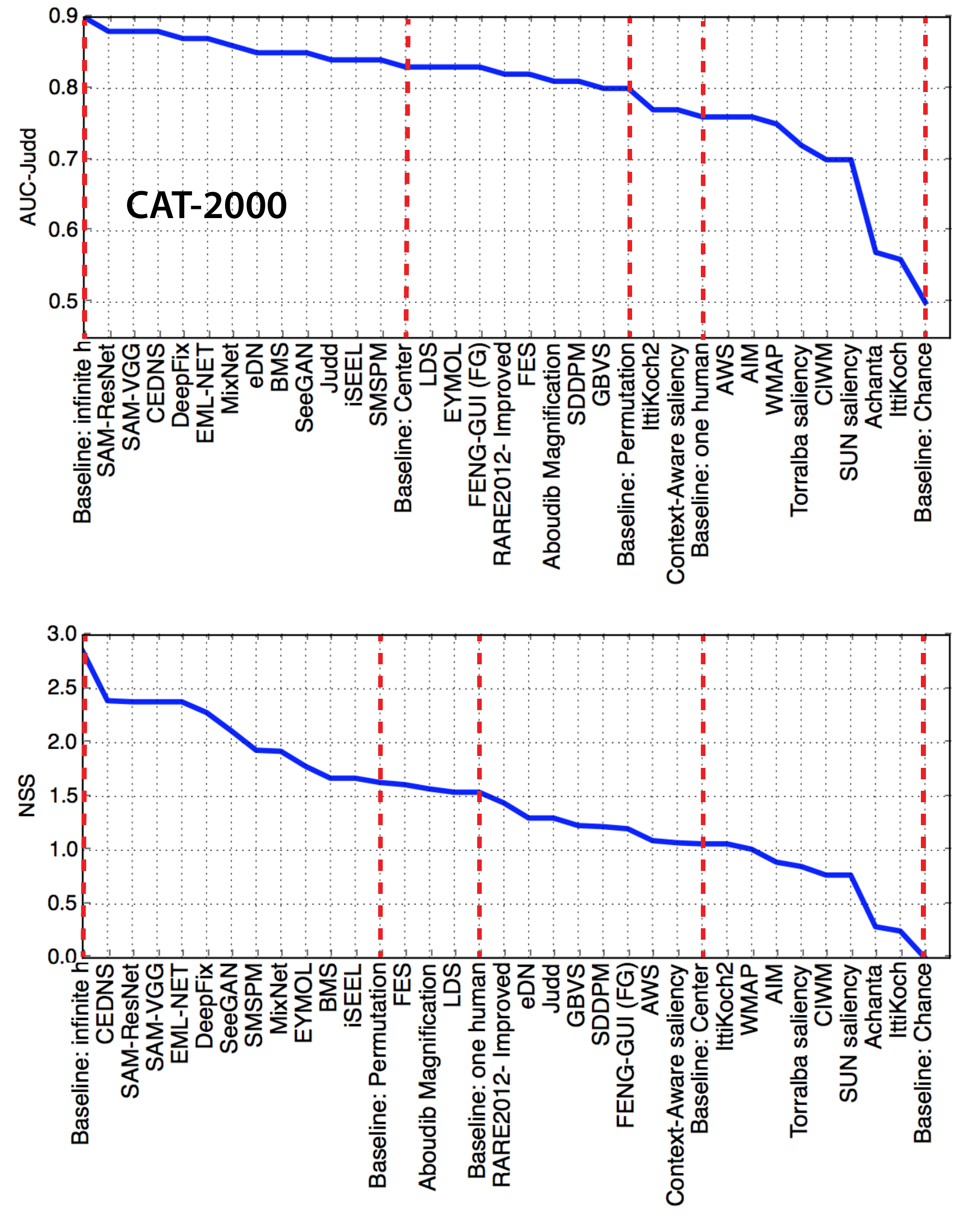} \\
	\caption{Performance of 30 saliency models and 5 baselines (marked with dashed red lines) for fixation prediction over the CAT2000 dataset. Models are sorted in terms of the AUC-Judd (top row) and NSS (bottom row) scores. Infinite h. stands for \textit{infinite humans} baseline.}
	\label{fig:catRes}
\end{figure}

Figure~\ref{fig:catRes} shows the results over the CAT2000 dataset. 
According to the AUC-Judd measure, SAM-ResNet, SAM-VGG, and CEDNS are tied in the top but still below the \textit{infinite humans} baseline. EML-Net, the winner on the MIT300 dataset, is ranked second here (tied with DeepFix). 
Switching to NSS, CEDNS performs slightly better than SAM-ResNet, SAM-VGG, and EML-Net. Among classic models, BMS~\cite{zhang2013saliency} and EYMOL~\cite{zanca2017variational} perform better than the others. Overall, models that do well over the MIT300 dataset perform well here as well.


There is a 3.5\% improvement from the best non-deep model to the best deep model in terms of AUC-Judd. 
The performance gap between the best non-deep model and the \textit{infinite humans} upper-bound is 5.55\% and shrinks to 2.22\% using the best deep learning model (based on the AUC-Judd measure). 


\begin{figure}
	\centering
    \includegraphics[width=.9\linewidth ]{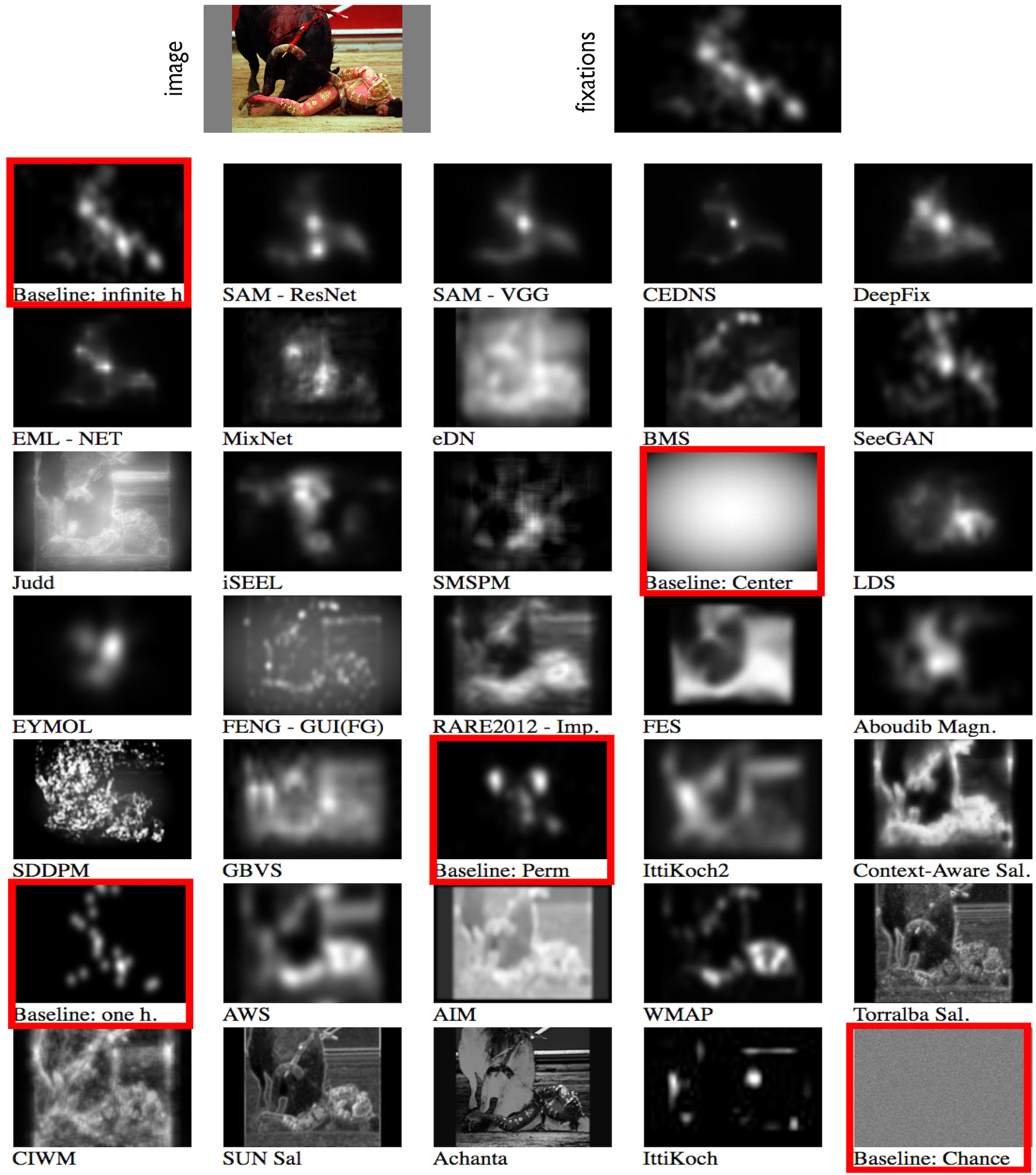} \\
	\caption{Saliency prediction maps of 35 models (including 5 baselines) on a sample image from the CAT2000 dataset (See the MIT benchmark webpage: \href{saliency.mit.edu}{saliency.mit.edu}). Red boxes illustrate baselines.}
	\label{fig:maps}
\end{figure}

A qualitative comparison of model predictions over a sample image from the CAT2000 dataset is presented in Figure~\ref{fig:maps}. It shows large differences in appearance of the maps generated by different models. At it can be seen, recent deep models generate saliency maps that are very similar to the ground truth fixation map on this image, better than their non-deep predecessors. 


\begin{figure}
	\centering
    \includegraphics[width=\linewidth ]{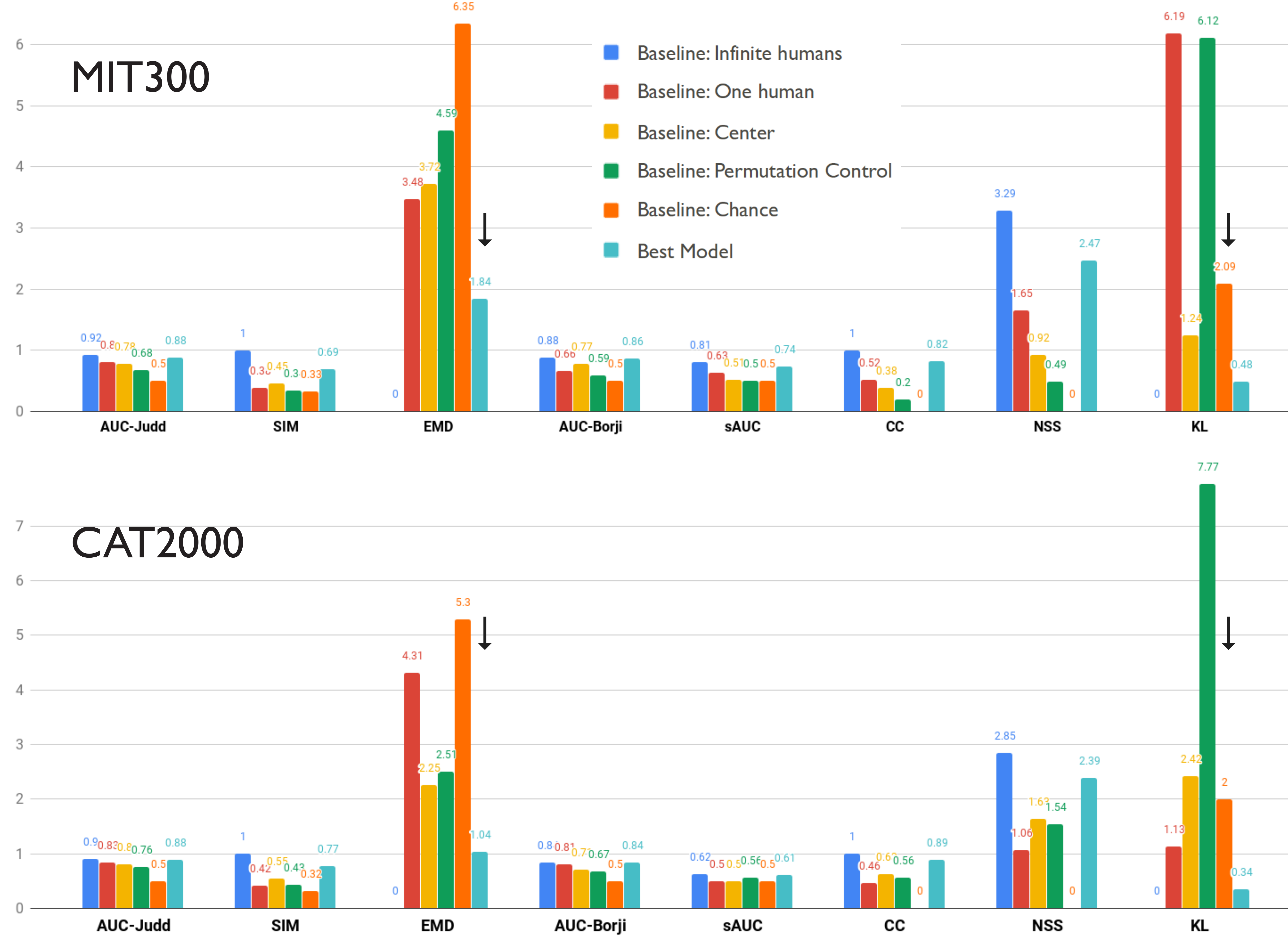} \\
\caption{Performance of five baselines and the best saliency model (can be different for each score) for predicting fixations over two eye movement datasets (MIT300 and CAT2000) using 8 scores. For EMD and KL, the lower the better (downward arrows). As it can be
seen, the best model often wins over the baselines except \textit{infinite humans} baseline, indicating a gap between the best models and humans in saliency prediction. The gap is wider for some scores (\eg NSS, EMD) and narrower using some other scores (\eg AUC-Borji, sAUC). This holds over both datasets. }
	\label{fig:charts}
\end{figure}

Overall, results show that new neural network based saliency models have created a large gap in performance relative to traditional saliency models from the pre deep learning era. However, they still fall short in performing at the level of humans on this task as is demonstrated in Figure~\ref{fig:charts}. Further, they suffer from several shortcomings that need to be addressed, and will be discussed next.

\subsection{Where do models fail?}
\label{missing}

\begin{figure*}[t]
	\centering
    \includegraphics[width=.8\linewidth]{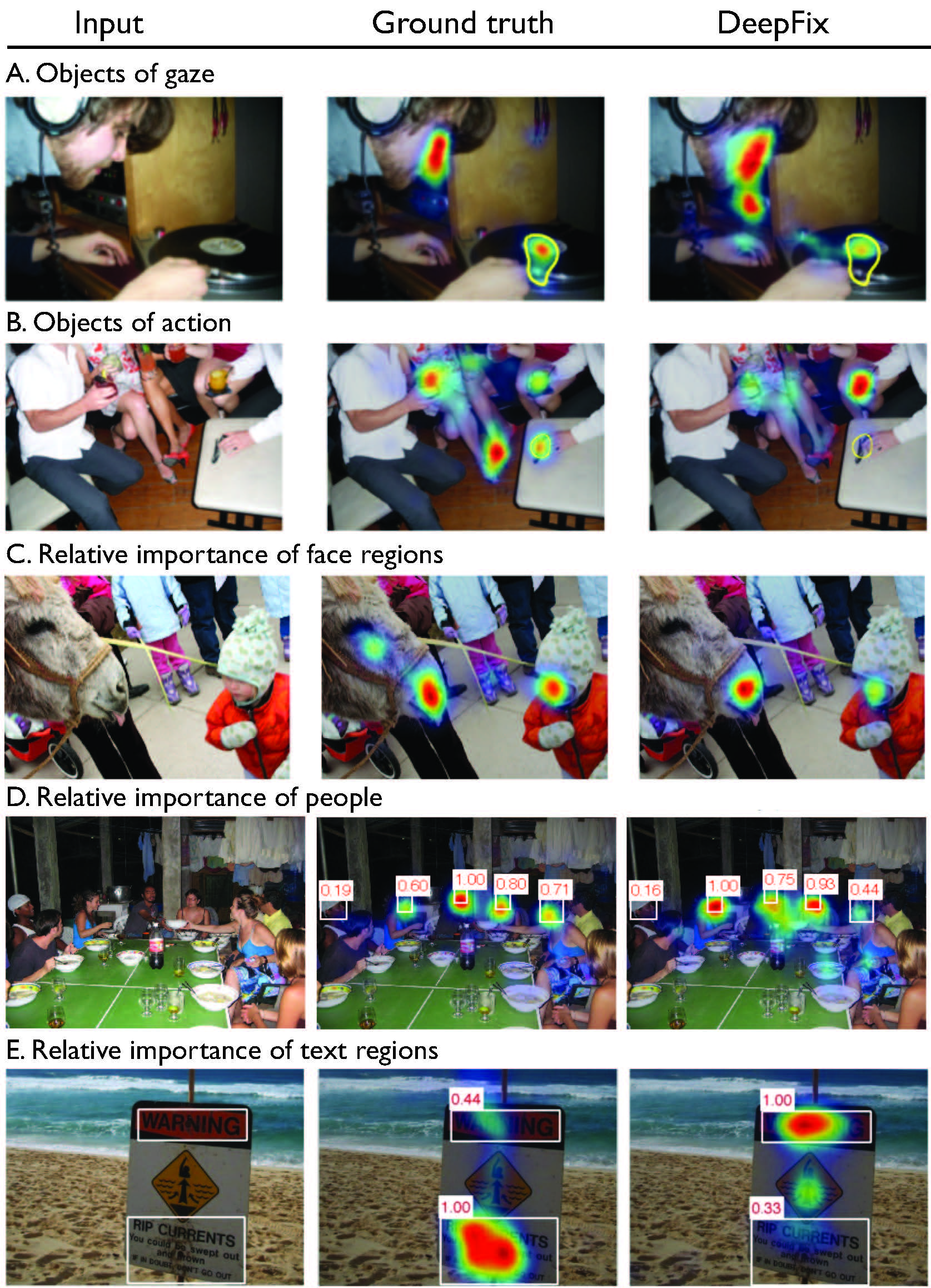} \\
	\caption{While deep learning models have shown impressive performance for saliency prediction, a finer-grained analysis shows that they miss key elements in images. 
	Some example stimuli for which models (here DeepFix model~\cite{kruthiventi2017deepfix}) under- or overestimate fixation locations due to gaze direction (A), or 
	locations of implied action or motion (B) are shown. 
Also, models fail to detect small faces or profile ones, and fail to assign correct relative importance to them (C). Cases for which models do not correctly assign relative importance to people (D), or text regions (E) in the scene are also shown. Please see text and~\cite{BylinskiiECCV2016} for further details.
}

   \vspace*{-5pt}
	\label{fig:failures}
\end{figure*}

As we saw above, deep learning models have shown impressive performance in saliency prediction. A deeper look, however, reveals that they continue to miss key elements in images (Figure~\ref{fig:failures}). 

In~\cite{BylinskiiECCV2016}, we investigated the state-of-the-art image saliency models using a fine-grained analysis on image types, image regions, etc. 
In a behavioral study, conducted via Amazon Mechanical Turk, workers were asked to label (1 out of 15 choices) image regions that fall on 
the top 5\% of the fixation heatmap. Analyzing the failures of models on those regions shows that the majority of the errors made by models are due to failures in accurately detecting parts of a person, faces, animals, text, objects of action, and gaze direction. These regions carry the greatest semantic importance in images (Figure~\ref{fig:failures}.A \& B). One way to ameliorate such errors is to train models on more instances of faces (\eg partial, blurry, small, or occluded faces, non-frontal views), more instances of text (different sizes and types), and animals. Saliency models may also need to be trained on different tasks, to learn to detect gaze and action and leverage this information for saliency. Moreover, saliency models need to reason about the relative importance of image regions, such as focusing on the most important person in the room or the most informative sign on the road (Figure~\ref{fig:failures}.C, D \& E). Interestingly, when we added the missing regions to models, performance improved drastically~\cite{BylinskiiECCV2016}. This has been corroborated by several other studies that showed augmented deep saliency models perform better than original models (\eg~\cite{nips15_recasens,gorji2017attentional,gorji2018going}).

\begin{figure}[t]
	\centering
    \includegraphics[width=\linewidth]{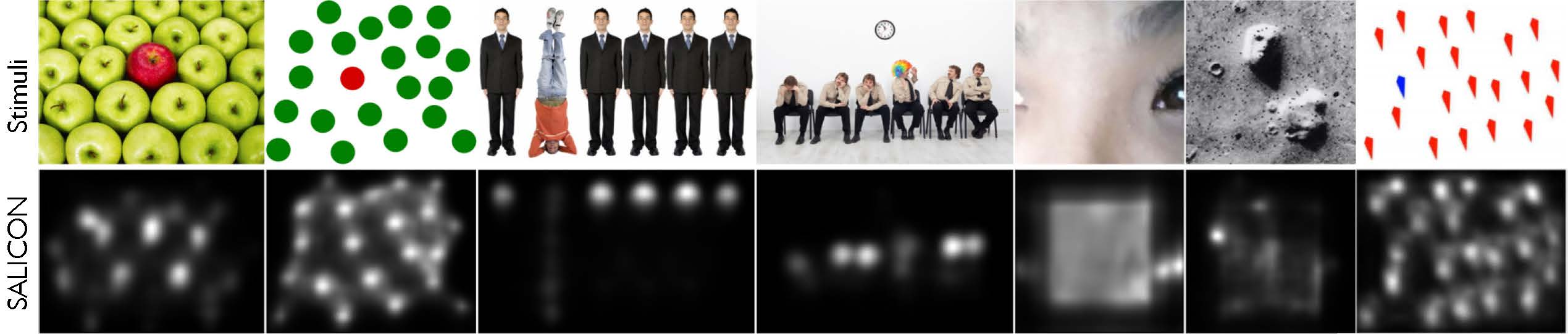}
	\caption{Example stimuli where deep saliency models produce counterintuitive results relative to models
based on feature contrast. 
Sometimes deep models neglect low-level image features (local contrast) and overweight the contribution of high-level features (\eg faces or text). Some of the failure cases might be due to the fact that deep saliency models compute absolute saliency, and not relative saliency (See Section 2.4). Although fixations are not recorded on these images, it is easy to predict where participants will typically fixate. See~\cite{rahman2015saliency} and~\cite{huang2015salicon} for more details. }
\label{fig:bruce}
\end{figure}

Previous research has also identified cases where deep saliency models produce counterintuitive results relative to models based on feature contrast. For example, Rahman and Bruce~\cite{rahman2015saliency} and Huang~\etal~\cite{huang2015salicon} showed that, unlike the classic Itti~\etal model, new saliency models fail to highlight odd items in pop-out psychological patterns (Figure~\ref{fig:bruce}).
 

\subsection{Discussion and outlook}
\label{conclusion}

\begin{figure}[t]
	\centering
    \includegraphics[width=\linewidth]{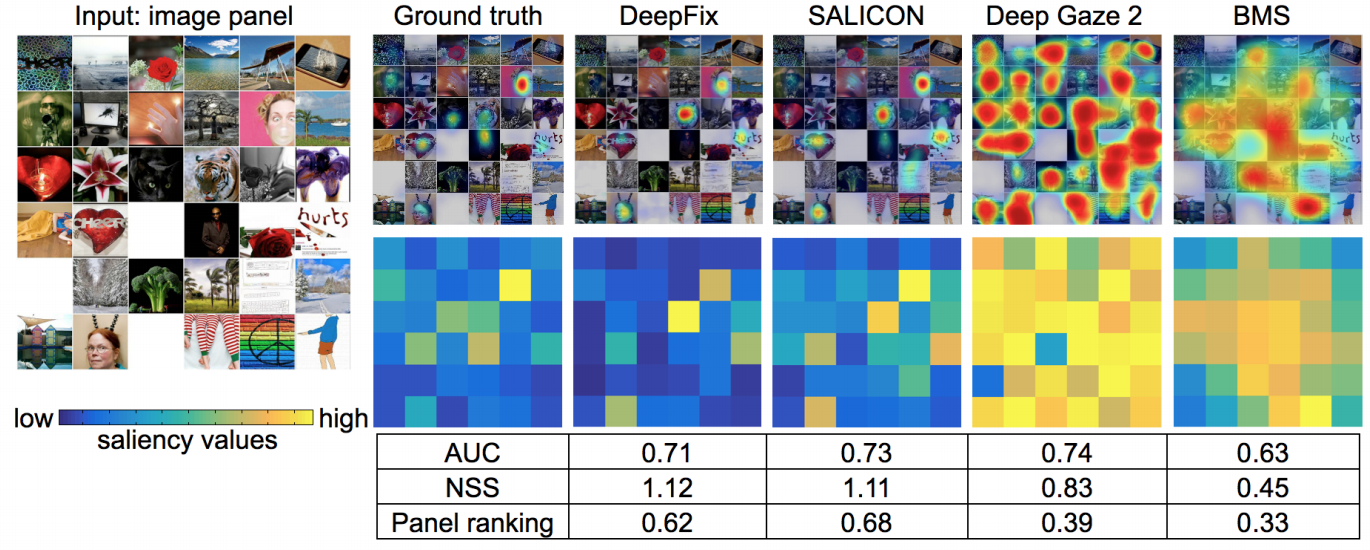}
	\caption{A finer-grained test proposed in~\cite{BylinskiiECCV2016} for determining how saliency models prioritize different sub-images in a panel, relative to each other. (left) A panel image from the MIT300 dataset. (right) The
saliency map predictions given the panel as an input image. The maximum response
of each saliency model on each subimage is visualized (as an importance matrix).
AUC and NSS scores are also computed for these saliency maps. The panel ranking is a measure of the correlation of values in the ground truth and predicted importance matrices. Figure from our work in~\cite{BylinskiiECCV2016}.}
	\label{fig:collage}
\end{figure}

Saliency prediction performance has improved dramatically in the last few years, in large part due to
deep supervised learning and large scale mouse click datasets. We also have a much better understanding of challenges pertaining to model evaluation than before. The new NN-based models are trained in a single end-to-end manner, combining feature extraction, feature integration, and saliency value prediction, and have created a large gap in performance relative to traditional saliency models. The success of these saliency prediction models suggests that the high-level image features encoded by deep networks (\eg sensitivity to faces, objects and text), as well as the ability of CNNs to capture global context are extremely useful for predicting fixation locations. Despite the immense recent progress, however, saliency prediction is far from being solved and there continues to be a big room for improvement. Some areas in which improvement can be made are discussed below:

\begin{itemize}


\item 
Saliency models based on deep learning are good face and text detectors, much better than their non-deep predecessors. The degree to which these models perform in face and text detection, compared to the state of the art face and text detectors, still remains to be determined.

\item
Even the best saliency models tend to place a disproportionate amount of importance on face regions, humans, and text even when they are not necessarily the most semantically interesting parts of the image. Saliency predictors will need to reason about the relative importance of image regions, such as focusing on the most important person in the room or the most informative sign on the road (which is image- and context-dependent). Similarly, in the presence of several text regions in the image, some high-level understanding of meanings is necessary to prioritize different text regions (\eg what is the warning sign about?)


\item There has been a lot of progress in understanding the representations learned by CNNs for scene and object recognition in recent years (\eg~\cite{zhou2014object}). Our understanding of what is learned by deep saliency models, however, is limited. The main questions here are 
how saliency computation emerges inside deep saliency architectures, and how the patterns in different network layers learned for saliency prediction differ from those patterns learned for object recognition? We have recently started to look into this (See He~\etal~\cite{ourVis}).


\item Fair saliency model comparison still remains an unsolved research problem today. Active research is ongoing to understand the pros and cons of the saliency measures (\eg~\cite{kummerer2018saliency}). Many of the current saliency methods compete closely with one another at the top of the existing benchmarks and performances vary in a narrow band (See Figures~\ref{fig:MitRes} \&~\ref{fig:catRes}). Also, as we saw in Section~\ref{sta}, some evaluation measures have begun to saturate and the produced rankings by different models are often inconsistent with each other. Thus, as the number of saliency models grows and score differences between models shrink, evaluation measures should be adjusted to a) elucidate differences between models and fixations (\eg by taking into account the relative importance of spatial and temporal regions), and b) mitigate sensitivity to map smoothing and center-bias. Complementary to measures, finer-grained stimuli such as image regions in a collection or in a panel (as in Figure~\ref{fig:collage} to measure how well models predict the relative importance of image content), psychophysical patterns (pop-out search arrays and natural oddball scenes), as well as transformed images can be used to further differentiate among models.


\item Collecting large scale data for constraining, training and evaluating attention models is crucial to progress. Bruce \etal~\cite{bruce2016deeper} pointed out that the manner in which data is selected, ground truth is created, and prediction error is measured (\eg loss function) is critical to model performance. Large scale click datasets~\cite{huang2015salicon,kim2017bubbleview} have been highly useful to train deep saliency models and to achieve high accuracy. However, clicks occasionally disagree with fixations. Thus, separating good clicks from noisy ones, can improve model training. Moreover, studying discrepancies between mouse movements and eye movements, collecting new types of image and video data, as well as refining available datasets can be rewarding.



\end{itemize}

In this review, we examined the recent progress in saliency prediction and proposed several avenues for future research. In spite of tremendous efforts and huge progress, there continues to be room for improvement in terms 
finer-grained analysis of deep saliency models, evaluation measures, datasets, annotation methods, cognitive studies, and new applications.

\section{Cross-References}

\noindent Hierarchical Models of the Visual System \\
\noindent Saliency in the Visual Cortex \\
\noindent Working Memory, Models of \\
\noindent Attentional Top-Down Modulation, Models of \\

\bibliography{ilab.bib}
\bibliographystyle{abbrv}

\end{document}